\documentclass[11pt]{article}

\usepackage[utf8]{inputenc}
\usepackage{times}
\usepackage[margin=1in]{geometry}

\usepackage{amsmath,amsfonts,bm}

\def\eqref#1{equation~\ref{#1}}

\def\1{\bm{1}}

\DeclareMathAlphabet{\mathsfit}{\encodingdefault}{\sfdefault}{m}{sl}
\SetMathAlphabet{\mathsfit}{bold}{\encodingdefault}{\sfdefault}{bx}{n}

\usepackage{hyperref}
\usepackage{url}
\usepackage{graphicx}
\usepackage{booktabs}
\usepackage{multirow}
\usepackage{caption}
\usepackage{tabularx}
\usepackage{natbib}
\usepackage{amsmath}
\usepackage{amssymb}

\hypersetup{
    colorlinks=true,
    linkcolor=black,
    citecolor=blue,
    urlcolor=blue
}

\begin{document}

\begin{center}

\rule{\textwidth}{2.5pt}

\vspace{0.18cm}

{\fontsize{20}{23}\selectfont\bfseries
ADTGUARD: ADAPTIVE DEPTH-TIME\\
SAFEGUARDS FOR STREAMING INFERENCE\\
IN LARGE LANGUAGE MODELS
\par}

\vspace{0.22cm}

\rule{\textwidth}{0.9pt}

\vspace{0.42cm}


{\fontsize{12}{15}\selectfont\bfseries
Chuming Wei\textsuperscript{1,$*$}
\hspace{0.9em}
Churui Yang\textsuperscript{1,$*$}
\hspace{0.9em}
Zihan ZHANG\textsuperscript{2}
\hspace{0.9em}
Ruien Li\textsuperscript{3}
\par}

\vspace{0.10cm}

{\fontsize{12}{15}\selectfont\bfseries
Wusheng zhang\textsuperscript{4}
\hspace{1.2em}
Yan Zhang\textsuperscript{4,$\dagger$}
\hspace{1.2em}
Guangwen Yang\textsuperscript{4}
\par}

\vspace{0.30cm}


{\fontsize{10.5}{14}\selectfont
\textsuperscript{1} Academy of Artificial Intelligence and Advanced Technology, Xi'an Jiaotong-Liverpool University\\
\textsuperscript{2} Faculty of Science and Engineering, University of Nottingham Ningbo China\\
\textsuperscript{3} Department of Computer Sciences, University of Wisconsin--Madison\\
\textsuperscript{4} Department of Computer Science and Technology, Tsinghua University
\par}

\end{center}

\vspace{0.32cm}

\begingroup
\renewcommand{\thefootnote}{\fnsymbol{footnote}}
\footnotetext[1]{Equal contribution.}
\footnotetext[2]{Correspondence to: Yan Zhang
\textless\texttt{yanzhang24@mails.tsinghua.edu.cn}\textgreater.}
\endgroup


\begin{abstract}
Large Language Models (LLMs) can generate fluent responses to diverse user requests, yet streaming may expose users to harmful content before generation is complete. Existing guardrails face two limitations in this setting. First, post-hoc moderation at response or chunk boundaries delays intervention, risking harmful exposure when content is released before verification. Second, even with token-level detection, coarse response-level supervision does not specify when harmful content begins, while fixed-layer readouts may miss relevant safety signals at other depths. To address these limitations, we introduce ADTGuard, which checks each token before release, progressively refines token-level training targets, and adapts layer selection to the current prefix. ADTGuard models risk across depth and time through Depth-Time Safety Dynamics (DTSD), which integrates safety evidence across layers at each token and uses it to guide risk accumulation over successive tokens. To adapt the choice of observation layers to the current prefix, we propose the Causal Depth Router (CDR). With only response-level labels available, we further develop Risk-Onset Target Refinement (RTR) to progressively construct token-level training targets for both components. Through extensive experiments across multiple safety benchmarks, we demonstrate that ADTGuard consistently achieves strong streaming detection performance, matching or surpassing state-of-the-art safeguards while requiring only 16M trainable parameters and maintaining competitive per-token latency.
\end{abstract}


\raggedbottom

\section{Introduction}

Large language models (LLMs) are increasingly deployed in interactive settings where
outputs are generated and delivered incrementally. Despite substantial progress in
alignment and content moderation, capable models can still be induced to generate
harmful or policy-violating content under adversarial or challenging inputs
\citep{andriushchenko2025adaptive,sharma2025constitutional}. This is particularly
consequential for autoregressive generation, where unsafe text may reach the user
before a response-level safeguard can evaluate the completed response
\citep{inan2023llamaguard,wang2025selfdefend,yu2026sentguard}. Effective streaming
safeguards must therefore assess risk as generation unfolds and enable intervention
before harmful content is exposed to the user.

Recent work has advanced safety monitoring from post-hoc detection to chunk-level
and ultimately per-token detection
\citep{xuan2025shieldhead,zhao2025qwen3guard,li2025kelp}.
However, finer temporal granularity does not make internal observation adaptive.
Most representation-based safeguards still rely on a fixed layer or predetermined
internal states
\citep{xuan2025shieldhead,jiao2026siren,cunningham2026constitutionalpp},
although safety information can vary substantially across model depth
\citep{li2025safetylayers,zhao2025harmrefusal}. A detector may therefore operate at
token granularity while still observing the model from a fixed depth. This raises a
complementary question: \emph{where within the model should safety evidence be read
as generation evolves?}

A second challenge is supervision granularity. Most publicly available safety
datasets provide response-level labels rather than token-level annotations of
harmful onset \citep{han2024wildguard,ji2023beavertails,ghosh2025aegis2}.
Obtaining such fine-grained token-level annotations manually is costly and
difficult to scale. Broadcasting a harmful response label to every decoding
position incorrectly marks early benign prefixes as unsafe
\citep{cunningham2026constitutionalpp,he2026segmentcoherence}, whereas manually
designed temporal priors impose a shared risk progression across responses whose
harmful content may emerge at different positions \citep{li2025kelp}. Thus,
token-level monitoring alone does not resolve the mismatch between coarse
response-level supervision and evolving prefix risk.

To address these challenges, we propose ADTGuard, a lightweight plug-in
safeguard that jointly adapts \emph{where} safety evidence is read and \emph{how}
token-level risk is learned. A Causal Depth Router (CDR) conditions on the
prompt and causal response prefix to select a sparse subset of model layers for each
response token. The routed representations are then modeled by
Depth-Time Safety Dynamics (DTSD), which integrates safety evidence across
selected depth positions and tracks its evolution over response tokens using
selective state-space modeling \citep{gu2024mamba}. Together, CDR and DTSD
characterize how safety evidence evolves across model depth and generation time as
the response unfolds, while keeping the underlying LLM frozen.

To recover fine-grained supervision from response-level labels, we introduce
Risk-Onset Target Refinement (RTR). RTR uses the detached risk trajectory cached
during the preceding training epoch to progressively construct refined token-level
soft targets \citep{laine2017temporalensembling,wang2021proselflc}, refining coarse
response labels toward temporally differentiated risk estimates. This enables
fine-grained risk learning without requiring explicit token-level onset annotations.
We further characterize the refinement as the solution to a weighted forward-KL
objective and use response-equal loss aggregation to prevent response length from
implicitly reweighting training examples.

Our contributions are summarized as follows:
\begin{enumerate}
    \item We propose \textbf{ADTGuard}, a lightweight streaming safeguard that
    adaptively selects safety-relevant internal observations as generation unfolds.
    Its \textbf{Causal Depth Router (CDR)} performs prefix-aware sparse routing over
    model depth, while \textbf{Depth-Time Safety Dynamics (DTSD)} jointly
    models safety evidence across depth and generation time over a frozen LLM.

    \item We introduce \textbf{Risk-Onset Target Refinement (RTR)}, which
    progressively derives fine-grained token targets from response-level
    supervision without requiring costly token-level annotations.

    \item We conduct extensive experiments across multiple safety benchmarks against
    representative response-level, representation-based, and streaming safeguards.
    ADTGuard achieves strong streaming detection performance while retaining a
    lightweight trainable footprint and competitive per-token latency.
\end{enumerate}
\section{Related Work}

\subsection{Streaming Safety Detection Based on Internal Representations}

Internal representations provide useful signals for safety monitoring, with prior
work identifying refusal-related directions, separating harmfulness from refusal
behavior, and revealing substantial variation in safety information across model
depth \citep{arditi2024refusal,zhao2025harmrefusal,li2025safetylayers}. Recent
safeguards increasingly exploit such signals at token level. ShieldHead performs
token-wise detection from hidden representations \citep{xuan2025shieldhead}, while
Disentangled Safety Adapters and HiddenGuard use specialized internal
representations for safety classification or intervention
\citep{krishna2026dsa,mei2026hiddenguard}. SIREN aggregates safety-relevant neurons
across layers \citep{jiao2026siren}, whereas Kelp, TrajGuard, Constitutional
Classifiers++, and Segment-Level Coherence model risk across model depth or
generation time
\citep{li2025kelp,liu2026trajguard,cunningham2026constitutionalpp,he2026segmentcoherence}.
However, their internal observation schemes are typically fixed after training
rather than adapting token by token to the evolving response prefix.

\subsection{Adaptive Depth and Sequential Modeling}

Adaptive computation methods dynamically vary how information is processed across
model depth. Depth-Adaptive Transformer and Confident Adaptive Language Modeling
adjust executed depth \citep{elbayad2020depthadaptive,schuster2022calm}, while
Mixture-of-Depths and Mixture-of-Recursions route tokens through different
computation depths \citep{raposo2024mod,bae2025mor}. Dynamic layer aggregation and
token--layer selective probes further learn input-dependent readouts across depth
\citep{dou2019dynamiclayer,meyoyan2026tokenlayer}. These methods, however, primarily
target general inference or classification rather than adaptive safety readout over
a frozen LLM whose full depth is executed.

Structured state-space models such as S4 and Mamba provide efficient sequential
modeling, with Mamba introducing input-dependent selective state updates
\citep{gu2022s4,gu2024mamba}. Related safety methods also model temporal risk
trajectories \citep{li2025kelp,cunningham2026constitutionalpp}, but depth and time
are typically treated separately. Consequently, temporal modeling does not
explicitly account for where safety evidence is read within the model or how
depth-wise observations interact with evolving prefix risk.

\subsection{Training Strategies for Token-Level Safety Detection}

Streaming detectors are often trained from response-level labels, creating a
mismatch with token-level predictions. Existing approaches use temporal aggregation
or regularization, including cumulative-maximum objectives, smoothing, and
window-based pooling
\citep{sharma2025constitutional,cunningham2026constitutionalpp,he2026segmentcoherence}.
ShieldHead and Kelp derive finer token-level signals using class prototypes or
temporal constraints \citep{xuan2025shieldhead,li2025kelp}, while SCM,
Qwen3Guard-Stream, and StreamGuard rely on sentence-level judgments, unsafe
boundaries, or continuation-based risk estimates
\citep{li2025scm,zhao2025qwen3guard,kavumba2026streamguard}. More broadly,
multiple-instance learning and self-training show that coarse labels can support
instance-level aggregation or progressively refined targets
\citep{ilse2018attentionmil,laine2017temporalensembling,wang2021proselflc}.
Nevertheless, fine-grained supervision often still depends on additional
assumptions, annotations, or rollout procedures, leaving open how to derive
temporally differentiated token-level supervision directly from response-level
labels.
\section{Methodology}
\label{sec:method}

\begin{figure*}[t]
    \centering
    \includegraphics[width=\textwidth]{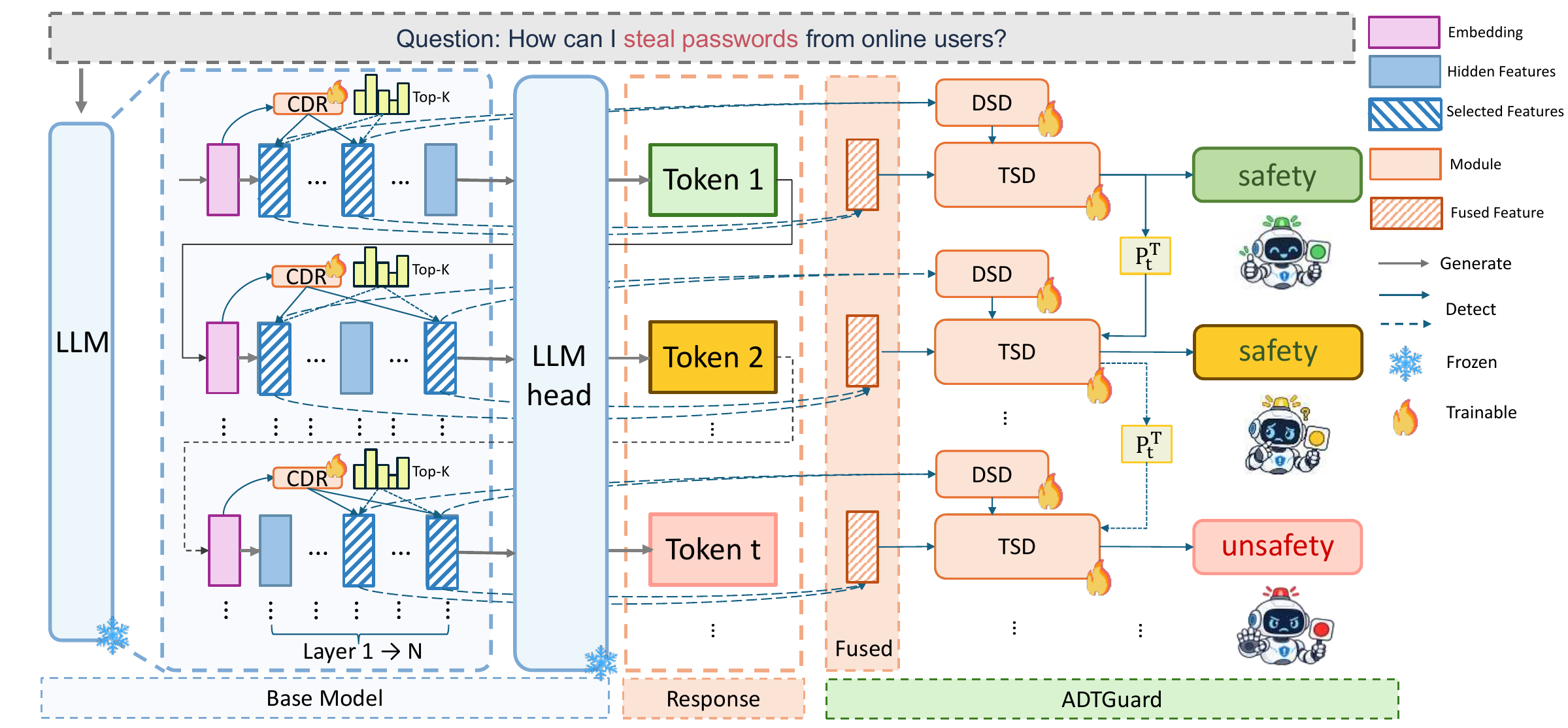}
    \caption{Overview of ADTGuard.
    At each generation step, CDR selects a prefix-dependent Top-$K$ set of layers
    and normalizes their routing weights. The selected features are individually
    reweighted and processed in depth order by DSD, while their weighted fusion is
    fed to TSD. The DSD output further conditions the TSD update, coupling depth-wise
    safety evidence with temporal risk evolution to produce a causal safety score for
    each response prefix over a frozen LLM.}
    \label{fig:adtguard_architecture}
\end{figure*}

We turn read-only hidden states of a frozen LLM into a streaming safeguard that
evaluates the harmfulness of each generated response prefix. Existing
representation-based safeguards typically monitor fixed layers
\citep{xuan2025shieldhead,jiao2026siren,cunningham2026constitutionalpp}, despite
substantial variation in safety-relevant information across model depth
\citep{li2025safetylayers,zhao2025harmrefusal}. Meanwhile, response-level safety
labels \citep{han2024wildguard,ji2023beavertails,ghosh2025aegis2} leave token-level
risk unspecified, leading detectors to rely on label broadcasting, hand-crafted
temporal priors, or additional boundary or rollout supervision
\citep{cunningham2026constitutionalpp,he2026segmentcoherence,zhao2025qwen3guard,kavumba2026streamguard}.
ADTGuard addresses these limitations through prefix-dependent layer selection,
depth-time risk modeling, and progressive token-target refinement.
As shown in Figure~\ref{fig:adtguard_architecture}, ADTGuard integrates these
components into a causal streaming detection pipeline over a frozen LLM. A frozen LLM with $L$ layers receives a prompt
$Q_{1:n}=(q_1,\ldots,q_n)$ and generates a response
$R_{1:m}=(r_1,\ldots,r_m)$. We denote the layer-$\ell$ hidden states of prompt
token $q_i$ and response token $r_t$ by
$\mathbf u_i^{(\ell)},\mathbf h_t^{(\ell)}\in\mathbb R^d$, respectively, where
$d$ is the backbone hidden dimension. All hidden states are accessed read-only.

\subsection{Causal Depth Router}
\label{sec:cdr}

\paragraph{Prompt initialization.}
To avoid repeatedly processing the prompt during decoding, we compute an initial
state $\mathbf s_0$ once before generation. We average each prompt token's hidden
states across all $L$ layers, apply the shared projection $\phi$, and pool over
prompt tokens:
\begin{equation}
    \mathbf s_0=\operatorname{Pool}\!\left(
    \left\{\phi\!\left(\frac{1}{L}\sum_{\ell=1}^{L}
    \mathbf u_i^{(\ell)}\right)\right\}_{i=1}^{n}\right).
    \label{eq:prompt_init}
\end{equation}
Here, $\phi:\mathbb R^d\rightarrow\mathbb R^{d_p}$ projects states to dimension
$d_p$, and $\operatorname{Pool}$ combines attention pooling with a learned mapping.
The resulting $\mathbf s_0\in\mathbb R^{d_p}$ initializes the temporal branch
and provides prompt context to the Router.

\paragraph{Prefix-dependent layer selection.}
We apply $\phi$ to the response hidden states $\mathbf h_t^{(0)}$ from the
embedding layer. Conditioned on $\mathbf s_0$, causal self-attention produces
contextualized Router states:
\begin{equation}
    [\mathbf c_0,\mathbf c_1,\ldots,\mathbf c_m]
    =
    \operatorname{CausalAttn}\!\left(
    [\psi(\mathbf s_0),
    \psi(\phi(\mathbf h_1^{(0)})),\ldots,
    \psi(\phi(\mathbf h_m^{(0)}))]
    \right).
    \label{eq:router_context}
\end{equation}
We discard $\mathbf c_0$ and use $\mathbf c_t$ as the Router context for
response position $t$. Here, $\psi$ maps inputs into the $d_r$-dimensional
Router space, $\mathbf W_q$ is a learned query projection, and we use sinusoidal
positional encodings \citep{vaswani2017attention}. Each layer has a learned depth
key $\mathbf k_\ell\in\mathbb R^{d_r}$. With routing budget $1\le K\le L$, CDR
scores all layers, selects the top $K$, and normalizes scores within the selected set:
\begin{align}
    e_{t,\ell}&=\frac{(\mathbf W_q\mathbf c_t)^\top\mathbf k_\ell}{\sqrt{d_r}},
    &\mathcal I_t&=\operatorname{TopK}_{\ell\in\{1,\ldots,L\}}(e_{t,\ell},K),
    \label{eq:cdr_score_topk}\\
    \alpha_{t,j}&=\frac{\exp(e_{t,\ell_{t,j}})}
    {\sum_{j'=1}^{K}\exp(e_{t,\ell_{t,j'}})},
    &\ell_{t,1}&<\cdots<\ell_{t,K}.
    \label{eq:cdr_weight}
\end{align}
The selected set $\mathcal I_t=\{\ell_{t,1},\ldots,\ell_{t,K}\}$ is ordered by
depth with weights $\alpha_{t,j}$. The frozen backbone still executes all layers;
CDR only determines which $K$ hidden states are read by the safeguard.

\subsection{Depth-Time Safety Dynamics}
\label{sec:dtsd}

Depth-Time Safety Dynamics (DTSD) contains two selective state-space branches:
Depth Safety Dynamics (DSD) integrates routed evidence across model depth, while
Temporal Safety Dynamics (TSD) tracks its evolution across response tokens
conditioned on the depth output. For
$\mathbf v_{t,j}=\phi(\mathbf h_t^{(\ell_{t,j})})$, we define
\begin{equation}
    \mathbf x^D_{t,j}=K\alpha_{t,j}\mathbf v_{t,j},
    \qquad
    \mathbf x^T_t=\sum_{j=1}^{K}\alpha_{t,j}\mathbf v_{t,j}.
    \label{eq:depth_time_inputs}
\end{equation}
The factor $K$ preserves the scale of each projected state under uniform routing.
Conceptually, letting $\mathbf x_t^D(u)$ and $\mathbf x^T(\tau)$ denote the
continuous counterparts of the above inputs, depth and temporal states
$\mathbf d_t(u),\mathbf s(\tau)\in\mathbb R^{d_p}$ evolve over normalized depth
$u\in[0,1]$ and generation progress $\tau$ as \citep{chen2018neuralode}
\begin{equation}
    \frac{\partial\mathbf d_t(u)}{\partial u}
    =F_D\!\left(\mathbf d_t(u),\mathbf x_t^D(u)\right),
    \quad
    \frac{\mathrm d\mathbf s(\tau)}{\mathrm d\tau}
    =F_T\!\left(\mathbf s(\tau),\mathbf x^T(\tau);\mathbf d(\tau)\right).
    \label{eq:continuous_dynamics}
\end{equation}
Here, $\mathbf d(\tau)$ denotes the corresponding depth summary; $F_D$ integrates
depth evidence, while $F_T$ evolves the temporal state conditioned on it.

\paragraph{Selective state-space realization.}
Both DSD and TSD use selective state-space recurrences with separate parameters
\citep{gu2022s4,gu2024mamba}. For input $\mathbf x_k$, a learned mapping $\Pi$
produces $\widehat{\boldsymbol\Delta}_k$, input coefficients $\mathbf B_k$,
and readout coefficients $\mathbf C_k$. For clarity, the recurrence below
is written for a single channel, where $x_k,\Delta_k,o_k,D$ are scalars and
$\mathbf q_k$ is the recurrent state:
\begin{alignat}{2}
(\widehat{\boldsymbol\Delta}_k,\mathbf B_k,\mathbf C_k)
&=\Pi(\mathbf x_k),
&\qquad
\boldsymbol\Delta_k
&=\operatorname{softplus}(\widehat{\boldsymbol\Delta}_k),
\label{eq:mamba_params}
\\
\bar{\mathbf A}_k
&=\exp(\Delta_k\mathbf A),
&
\bar{\mathbf B}_k
&=\Delta_k\mathbf B_k,
\label{eq:mamba_discretization}
\\
\mathbf q_k
&=\bar{\mathbf A}_k\mathbf q_{k-1}+\bar{\mathbf B}_k x_k,
&
o_k
&=\mathbf C_k\mathbf q_k+D x_k.
\label{eq:mamba_recurrence}
\end{alignat}
Here, $k$ indexes selected layers or response positions, $\mathbf A$ is a learned
diagonal transition matrix, and $D$ provides a direct input-to-output mapping.
The step size controls the rate of state evolution.

\paragraph{Depth Safety Dynamics.}
To account for uneven spacing between selected layers
\citep{rubanova2019latentode}, we scale the depth step size by their normalized
gaps. With $u_{t,j}=\ell_{t,j}/L$ and $u_{t,0}=0$,
\begin{equation}
    g_{t,j}=K(u_{t,j}-u_{t,j-1}),
    \qquad
    \boldsymbol\Delta^D_{t,j}=g_{t,j}\operatorname{softplus}
    (\widehat{\boldsymbol\Delta}^D_{t,j}).
    \label{eq:depth_gap_delta}
\end{equation}
The factor $K$ expresses each gap relative to the uniform spacing $1/K$.
The depth branch processes $\mathbf x^D_{t,1:K}$ in depth order, initializes its
recurrent state to zero for each response token, and uses the final selected-layer
output as $\mathbf d_t$.

\paragraph{Temporal Safety Dynamics.}
The temporal branch is initialized from $\mathbf s_0$ and carries its recurrent
state across response positions. At position $t$, it receives $\mathbf x_t^T$,
while the depth output modulates its step size:
\begin{equation}
    \boldsymbol\Delta^T_t=\operatorname{softplus}\!\left(
    \widehat{\boldsymbol\Delta}^T_t+\gamma(\mathbf d_t)\right).
    \label{eq:time_delta}
\end{equation}
The learned mapping $\gamma$ adjusts the base step size according to the current
depth evidence, coupling depth-wise safety information to temporal risk evolution.

\paragraph{Risk prediction.}
Each branch $b\in\{D,T\}$ produces benign and harmful logits
$\mathbf z_t^b=(z_{t,0}^b,z_{t,1}^b)$:
\begin{equation}
    a_t^b=z_{t,1}^b-z_{t,0}^b,
    \qquad
    p_t^b=\sigma(a_t^b)=\operatorname{softmax}(\mathbf z_t^b)_1.
    \label{eq:risk_head}
\end{equation}
The temporal prediction $p_t=p_t^T$ is the final risk score for the response
prefix up to position $t$.

\subsection{Risk-Onset Target Refinement}
\label{sec:rtr}

Response-level labels indicate overall harmfulness but leave token-level risk
unspecified. Risk-Onset Target Refinement (RTR) progressively derives token-level
targets from detached risk trajectories cached during the preceding epoch, avoiding
a separate teacher or additional inference pass
\citep{laine2017temporalensembling,wang2021proselflc}.
For a harmful response $i$ with $y_i=1$, the onset distribution
$\rho_{i,t}^{(e-1)}$ is obtained by normalizing the positive increments
$[p_{i,t}^{(e-1)}-p_{i,t-1}^{(e-1)}]_+$ of the cached trajectory over valid
positions, with cumulative profile
\begin{equation}
    c_{i,t}^{(e-1)}
    =\sum_{\tau=1}^{t}\rho_{i,\tau}^{(e-1)}.
    \label{eq:rtr_profile}
\end{equation}
With epoch-dependent weight $\beta_e\in[0,1]$, RTR refines the target as
\begin{equation}
    q_{i,t}^{(e)}
    =\beta_e+(1-\beta_e)c_{i,t}^{(e-1)},
    \qquad y_i=1,
    \label{eq:rtr_target}
\end{equation}
while benign responses use $c_{i,t}^{(e-1)}=q_{i,t}^{(e)}=0$. The first epoch
uses the response-level label throughout; decreasing $\beta_e$ progressively
shifts supervision toward the model-derived onset profile.

\paragraph{Forward-KL characterization of RTR.}
For $\operatorname{Ber}(a)$ denoting a Bernoulli distribution, response label
$y\in\{0,1\}$, and fixed profile $c\in[0,1]$ with $c=0$ for benign responses,
\begin{equation}
    \mathcal J(q)
    =
    \beta D_{\mathrm{KL}}\!\left(
    \operatorname{Ber}(y)\,\|\,\operatorname{Ber}(q)\right)
    +(1-\beta)D_{\mathrm{KL}}\!\left(
    \operatorname{Ber}(c)\,\|\,\operatorname{Ber}(q)\right)
\end{equation}
has the unique minimizer
\begin{equation}
    q^\star=\beta y+(1-\beta)c.
    \label{eq:rtr_kl_solution}
\end{equation}
Thus, conditioned on a fixed model-derived onset profile, the RTR target is the
exact weighted forward-KL optimum between response-level supervision and the
onset profile; see Appendix~\ref{app:rtr_kl_proof}.

\subsection{Training Objective and Streaming Inference}
\label{sec:objective}

We train both branches with RTR targets, averaging binary cross-entropy within
each response and then across the batch so that responses receive equal weight:
\begin{equation}
    \mathcal L_{\mathrm{RE}}^b=
    \frac{1}{B}\sum_{i=1}^{B}\frac{1}{m_i}\sum_{t=1}^{m_i}
    \operatorname{BCEWithLogits}(a_{i,t}^b,q_{i,t}^{(e)}),
    \qquad b\in\{D,T\}.
    \label{eq:response_equal}
\end{equation}
Here, $B$ is the batch size and $m_i$ counts valid response tokens, excluding
padding and terminal EOS. To encourage nondecreasing temporal predictions, we
penalize risk decreases over valid adjacent token pairs:
\begin{equation}
    \mathcal L_{\mathrm{mono}}=
    \frac{\sum_{i=1}^{B}\sum_{t=2}^{m_i}\max(p^T_{i,t-1}-p^T_{i,t},0)}
    {\max\{1,\sum_{i=1}^{B}(m_i-1)\}}.
    \label{eq:mono}
\end{equation}
The total loss is
\begin{equation}
    \mathcal L=\lambda_T\mathcal L_{\mathrm{RE}}^T
    +\lambda_D\mathcal L_{\mathrm{RE}}^D
    +\lambda_M\mathcal L_{\mathrm{mono}},
    \label{eq:total_loss}
\end{equation}
where $\lambda_T,\lambda_D,\lambda_M$ weight the three terms.
Training processes response positions in parallel while preserving causality.
During streaming inference, the LLM and Router KV caches and the recurrent TSD
state are reused across decoding steps, avoiding recomputation of earlier tokens.
Each risk prediction therefore depends only on the prompt and currently available
response prefix.
\section{Experiments}
\label{sec:experiments}

\subsection{Experimental Setup}
\label{sec:experimental-setup}

We use four NVIDIA A800 GPUs (80~GB each) for the default training setup.
Qwen3-8B Base~\citep{yang2025qwen3} serves as the primary backbone, with
additional experiments on Llama-3.1-8B Base~\citep{grattafiori2024llama3}
and Gemma-7B Base~\citep{gemmateam2024gemma}. During training, the backbone
remains frozen and only the attached ADTGuard module is updated, with
approximately 15.6M trainable parameters. For each backbone--dataset pair,
a separate detector is trained for 5 epochs with a global batch size of 8
and a maximum combined prompt--response length of 4,096 tokens. Training
uses AdamW with a learning rate of $5\times10^{-5}$, weight decay of 0.1,
a warmup ratio of 0.05, and a cosine learning-rate schedule. Unless otherwise
specified, the same ADTGuard configuration is fixed across all
backbone--dataset settings; the analyses in Section~4.5 are post-hoc
sensitivity studies and do not affect the reported main results. Further
implementation details are provided in
Appendix~\ref{app:adtguard-configuration}.

\noindent\textbf{Datasets.}
We evaluate on five safety benchmarks: SGB-S-Eval and SGB-WildGuard from
StreamGuardBench (SGB)~\citep{li2025kelp}, AEGIS2.0~\citep{ghosh2025aegis2},
BeaverTails~\citep{ji2023beavertails}, and WildGuard~\citep{han2024wildguard}.
Within SGB, SGB-S-Eval and SGB-WildGuard contain generated and annotated
responses to prompts drawn from S-Eval and WildGuard, respectively, whereas
WildGuard uses the original public responses and safety annotations.

\noindent\textbf{Baselines.}
We compare ADTGuard against two groups of safeguards. (i) Dedicated moderation models include Qwen3Guard-Gen-8B, Qwen3Guard-Stream-8B~\citep{zhao2025qwen3guard}, Llama-Guard-4-12B~\citep{meta2025llamaguard4}, and SentGuard-4B~\citep{yu2026sentguard}, all adapted to the corresponding training split using LoRA. (ii) Representation-based safeguards are evaluated on the same backbone as ADTGuard. SIREN~\citep{jiao2026siren} aggregates safety-relevant neurons across layers, while Kelp~\citep{li2025kelp} incorporates recurrent latent dynamics to track risk over time.

\noindent\textbf{Metrics.}
Safety evaluation should capture both the correctness of the final response judgment and the ability to identify risk as generation unfolds. Therefore, we use unsafe-class F1 as the primary metric, treating unsafe responses as the positive class, and report two variants: (i) \emph{Response F1}, computed from the prediction at the final response token; and (ii) \emph{Streaming F1}, where a response is considered unsafe if any evaluated prefix is detected as unsafe. Detailed evaluation protocols are provided in Appendix~\ref{app:evaluation-protocols}.

\subsection{Main Results}
\label{sec:main-comparison}

Table~\ref{tab:representation-comparison} compares representation-based safeguards on the same backbone. Table~\ref{tab:dedicated-comparison} compares dedicated moderation models.

\noindent\textbf{Comparison with representation-based safeguards.}
As shown in Table~\ref{tab:representation-comparison}, ADTGuard achieves the best Streaming F1 in 14 of 15 backbone--dataset settings and the best Response F1 in 14 of 15 settings. With Qwen3-8B, ADTGuard reaches Streaming F1 scores of 93.57 on SGB-S-Eval and 85.87 on SGB-WildGuard, outperforming the strongest baselines by 5.24 and 25.75 points, respectively. Similar gains are observed across Llama-3.1-8B and Gemma-7B, indicating that the improvement is not tied to a specific backbone. These results suggest that adaptive layer selection and joint depth-time modeling provide stronger streaming safety representations than fixed readout schemes. Notably, ADTGuard's Streaming F1 also remains close to its Response F1 across most settings, indicating consistent risk detection as generation unfolds.

\noindent\textbf{Comparison with dedicated moderation models.}
As shown in Table~\ref{tab:dedicated-comparison}, we further compare ADTGuard with standalone dedicated moderation models. In this comparison, ADTGuard uses Qwen3-8B Base and reuses its internal representations, requiring only an additional 15.6M-parameter detection module, whereas the dedicated guards operate as separate 4B--12B moderation models. Despite this difference in deployment paradigm, ADTGuard achieves the highest Streaming F1 on four of the five datasets, with the largest improvement of 5.05 points over Qwen3Guard-Stream-8B on SGB-WildGuard. These results show that leveraging the protected model's internal representations can provide competitive streaming detection without deploying an additional large-scale moderation model.

\begin{table}[!ht]
\centering
\caption{
Comparison on the same backbone.
F1 comparison among representation-based safeguards, including SIREN, Kelp, and ADTGuard, using Qwen3-8B Base, Llama-3.1-8B Base, and Gemma-7B Base as shared backbones. Response / Streaming unsafe-class F1 (\%) are reported. Bold and underline denote the best and second-best results, respectively, for each metric and dataset within the same backbone.
}
\label{tab:representation-comparison}
\begingroup
\fontsize{8}{10}\selectfont
\setlength{\tabcolsep}{3.0pt}
\renewcommand{\arraystretch}{1.08}
\begin{tabular*}{\linewidth}{@{\extracolsep{\fill}}lccccc@{}}
\toprule
Method & SGB-S-Eval & SGB-WildGuard & AEGIS2.0 & BeaverTails & Public WildGuard \\
\midrule
\multicolumn{6}{l}{\textit{Qwen3-8B Base}} \\
SIREN (in-domain) & \underline{92.22} / 79.84 & \underline{78.90} / 55.93 & \underline{84.40} / \underline{83.31} & \underline{87.02} / 82.99 & \underline{71.91} / \underline{60.08} \\
SIREN (official) & 55.36 / 56.91 & 70.32 / 52.31 & 82.84 / 78.75 & 85.67 / 82.16 & 65.71 / 53.54 \\
Kelp & 88.05 / \underline{88.33} & 53.59 / \underline{60.12} & 48.74 / 49.82 & 83.27 / \underline{83.20} & 55.04 / 57.52 \\
ADTGuard & \textbf{93.67} / \textbf{93.57} & \textbf{83.94} / \textbf{85.87} & \textbf{85.79} / \textbf{86.84} & \textbf{87.33} / \textbf{87.39} & \textbf{74.76} / \textbf{73.86} \\
\midrule
\multicolumn{6}{l}{\textit{Llama-3.1-8B Base}} \\
SIREN (in-domain) & \underline{91.10} / \underline{90.56} & \underline{81.12} / 49.56 & \textbf{86.75} / \textbf{85.85} & \underline{86.70} / 82.57 & \underline{69.38} / 59.26 \\
SIREN (official) & 51.93 / 58.58 & 67.10 / 51.53 & 81.71 / 79.05 & 85.96 / 82.29 & 64.05 / 52.91 \\
Kelp & 88.16 / 89.34 & 78.06 / \underline{72.95} & 80.69 / 82.31 & 84.98 / \underline{85.18} & 66.93 / \underline{73.05} \\
ADTGuard & \textbf{93.21} / \textbf{93.10} & \textbf{83.36} / \textbf{83.59} & \underline{84.14} / \underline{85.32} & \textbf{87.15} / \textbf{87.07} & \textbf{74.25} / \textbf{74.23} \\
\midrule
\multicolumn{6}{l}{\textit{Gemma-7B Base}} \\
SIREN (in-domain) & 78.74 / 69.54 & \underline{77.56} / 61.17 & 78.95 / 79.23 & 75.77 / 76.09 & \underline{68.35} / 38.62 \\
SIREN (official) & 53.87 / 57.47 & 64.72 / 59.61 & 78.64 / 78.10 & 66.55 / 66.78 & 61.50 / 56.23 \\
Kelp & \underline{86.39} / \underline{86.99} & 72.60 / \underline{70.11} & \underline{79.08} / \underline{81.61} & \underline{83.56} / \underline{83.67} & 65.05 / \underline{71.36} \\
ADTGuard & \textbf{93.81} / \textbf{93.51} & \textbf{81.98} / \textbf{82.45} & \textbf{84.75} / \textbf{86.15} & \textbf{87.20} / \textbf{87.50} & \textbf{72.65} / \textbf{74.72} \\
\bottomrule
\end{tabular*}
\endgroup
\vspace{3pt}
\parbox{\linewidth}{\footnotesize
SIREN (in-domain): trained separately on the corresponding benchmark training split.\\
SIREN (official): trained on the seven official datasets used by SIREN.
}
\end{table}

\begin{table}[!ht]
\centering
\caption{
Comparison with dedicated moderation models.
F1 comparison with standalone moderation models, including Qwen3Guard-Gen-8B, Qwen3Guard-Stream-8B, Llama-Guard-4-12B, and SentGuard-4B. ADTGuard uses Qwen3-8B Base as the protected backbone. Response / Streaming unsafe-class F1 are reported. Bold and underline denote the best and second-best results, respectively, for each metric and dataset.
}
\label{tab:dedicated-comparison}
\begingroup
\fontsize{8}{10}\selectfont
\setlength{\tabcolsep}{1.6pt}
\renewcommand{\arraystretch}{1.12}
\begin{tabular*}{\linewidth}{@{\extracolsep{\fill}}lccccc@{}}
\toprule
Model & SGB-S-Eval & SGB-WildGuard & AEGIS2.0 & BeaverTails & Public WildGuard \\
\midrule
Qwen3Guard-Gen-8B & \underline{0.9341} / 0.8806 & \textbf{0.8592} / 0.7903 & \underline{0.8645} / --- & 0.8723 / --- & \textbf{0.7892} / --- \\
Qwen3Guard-Stream-8B & 0.9338 / \underline{0.9278} & 0.8456 / \underline{0.8082} & \textbf{0.8752} / \underline{0.8625} & \textbf{0.8754} / \underline{0.8673} & \underline{0.7559} / \textbf{0.7828} \\
Llama-Guard-4-12B & 0.9242 / 0.8546 & 0.8419 / 0.7429 & 0.8340 / --- & 0.8623 / --- & 0.7403 / --- \\
SentGuard-4B & 0.9249 / 0.8294 & \underline{0.8546} / 0.7486 & 0.8311 / 0.8500 & 0.8592 / 0.8620 & 0.7109 / \underline{0.7559} \\
\midrule
ADTGuard & \textbf{0.9367} / \textbf{0.9357} & 0.8394 / \textbf{0.8587} & 0.8579 / \textbf{0.8684} & \underline{0.8733} / \textbf{0.8739} & 0.7476 / 0.7386 \\
\bottomrule
\end{tabular*}
\endgroup
\vspace{3pt}
\parbox{\linewidth}{\footnotesize
The four LoRA-adapted guards check every 10 tokens on SGB. On public datasets, Qwen3Guard-Stream checks tokens and SentGuard checks cumulative sentences. ``---'' denotes unreported results.
}
\end{table}

\subsection{Ablation Study}
\label{sec:ablation-study}

We conduct all ablations on SGB-S-Eval with Qwen3-8B Base, using the same data split, optimizer, five-epoch training budget, and random seed (42) throughout. All ablations report the final checkpoint at a threshold of 0.5.

\begin{table}[!ht]
\centering
\caption{
Component ablation.
Starting from an approximately parameter-matched token-wise MLP baseline, we progressively introduce CDR, replace the MLP with DTSD, and then add RTR and Mono on SGB-S-Eval with Qwen3-8B Base. Final-epoch unsafe-class F1 (\%) is reported. Bold and underline denote the best and second-best results, respectively.
}
\label{tab:component-ablation}
\begingroup\footnotesize
\setlength{\tabcolsep}{5pt}
\begin{tabular*}{0.70\linewidth}{@{\extracolsep{\fill}}lrr@{}}
\toprule
Configuration & Response & Streaming \\
\midrule
Baseline & 87.31 & 86.05 \\
\quad + CDR & 88.06 & 86.09 \\
\quad + DTSD & 92.83 & 87.51 \\
\quad + RTR & \underline{93.58} & \textbf{93.58} \\
\quad + Mono (ours) & \textbf{93.67} & \underline{93.57} \\
\bottomrule
\end{tabular*}
\endgroup
\end{table}

\noindent\textbf{Component Analysis.}
We progressively introduce CDR, DTSD, RTR, and Mono to an approximately parameter-matched token-wise MLP baseline to examine the contribution of each component. As shown in Table~\ref{tab:component-ablation}, adding CDR alone yields only a modest improvement, whereas combining CDR with DTSD increases Response F1 from 88.06 to 92.83 and Streaming F1 from 86.09 to 87.51. This suggests that adaptive layer selection becomes substantially more effective when the routed representations are further integrated through depth-time dynamics, allowing the detector to aggregate safety evidence across both selected model depths and successive response positions. RTR then produces the largest improvement in streaming performance, increasing Streaming F1 from 87.51 to 93.58 while also improving Response F1 to 93.58. This result highlights the importance of progressively refined token-level targets, which reduce the mismatch between coarse response-level supervision and the evolving risk of intermediate prefixes. Finally, the monotonicity regularizer penalizes decreases in temporal risk, encouraging a more consistent risk trajectory across generation. It slightly improves Response F1 from 93.58 to 93.67 while keeping Streaming F1 essentially unchanged, indicating that monotonic regularization mainly stabilizes the learned risk evolution without sacrificing streaming detection performance.

\subsection{Latency Evaluation}
\label{sec:latency-evaluation}

ADTGuard enables token-by-token safety monitoring with modest inference
overhead. We evaluate Qwen3-8B Base on a single NVIDIA A800 GPU (80~GB)
with batch size~1, retaining all token-level checks without early
termination. With a 1,000-token prompt and 1,024 output tokens, ADTGuard
adds 2.33~ms of amortized latency per output token, increasing total
request time from 19.32~s to 21.71~s (+12.36\%). At the same output
length, the added cost remains within 2.31--2.35~ms/token across prompt
lengths of 128--2,048 tokens. Together with the detection results above,
these measurements highlight ADTGuard's effective balance between
streaming detection performance and inference efficiency. Full latency
results and evaluation details are provided in
Appendix~\ref{app:latency-evaluation}.

\subsection{Hyperparameter Analysis}
\label{sec:hyperparameter-analysis}

We conduct a controlled hyperparameter analysis on SGB-S-Eval with Qwen3-8B Base. All experiments follow the same training configuration as the default ADTGuard setting, with only the hyperparameter under study varied in each comparison. Specifically, we examine two key choices: the routing budget $k$, which determines the number of selected observation layers, and the RTR schedule $\beta_{1:5}$, which controls the transition from response-level supervision to model-derived token-level targets. For the routing study, we vary $k$ from 1 to 9 while keeping the default RTR schedule fixed. For the target-refinement study, we fix $k=4$ and compare different RTR schedules over the five training epochs.

\begin{table}[!ht]
\centering
\caption{
Hyperparameter analysis.
Routing budget and target-refinement schedule on SGB-S-Eval with Qwen3-8B Base. Final-epoch unsafe-class F1 (\%) is reported. Panel (a) varies only the routing budget $k$, while panel (b) fixes $k=4$ and varies the five epoch-wise $\beta$ values. FP/FN counts refer to Streaming. Bold and underline denote the best and second-best results, respectively, within each panel and metric.
}
\label{tab:hyperparameter-ablation}
\footnotesize
\textbf{(a) Number of observation layers}\\[3pt]
\setlength{\tabcolsep}{3.4pt}
\begin{tabular*}{\linewidth}{@{\extracolsep{\fill}}lrrrrrrrrr@{}}
\toprule
$k$ & 1 & 2 & 3 & 4 & 5 & 6 & 7 & 8 & 9 \\
\midrule
Response & 91.30 & \underline{93.50} & 92.94 & \textbf{93.67} & 93.25 & 93.41 & 93.12 & 93.30 & 93.13 \\
Streaming & 91.87 & 93.21 & 92.95 & \textbf{93.57} & \underline{93.51} & 93.49 & 93.41 & 93.31 & 93.23 \\
\bottomrule
\end{tabular*}
\vspace{8pt}

\textbf{(b) Soft-target schedule}\\[3pt]
\begin{tabular*}{\linewidth}{@{\extracolsep{\fill}}llrrr@{}}
\toprule
Schedule & $\beta_{1:5}$ & Response & Streaming & FP / FN \\
\midrule
Hard targets & $[1,1,1,1,1]$ & 92.66 & 87.68 & 137 / 10 \\
Immediate & $[1,0,0,0,0]$ & 93.50 & 93.10 & 40 / 34 \\
Linear & $[1,\frac{2}{3},\frac{1}{3},0,0]$ & \underline{93.61} & 93.41 & 41 / 30 \\
Cosine & $[1,0.75,0.25,0,0]$ & 93.51 & \textbf{93.58} & 39 / 30 \\
Default & $[1,0.5,0.25,0,0]$ & \textbf{93.67} & \underline{93.57} & 38 / 31 \\
\bottomrule
\end{tabular*}
\end{table}

\noindent\textbf{Routing Budget.}
For $k=1$--$9$, Response and Streaming F1 remain consistently high, with the best overall performance obtained at $k=4$ (Table~\ref{tab:hyperparameter-ablation}(a)). In particular, $k=1$ yields noticeably lower scores than the multi-layer settings, suggesting that relying on a single observation layer may provide insufficient coverage of the distributed safety evidence across model depth. Increasing $k$ to 4 improves both Response and Streaming F1, indicating that a small set of adaptively selected layers is sufficient to capture complementary safety signals. Beyond $k=4$, however, larger routing budgets bring no further gains and instead produce small fluctuations in performance. Since increasing $k$ also requires processing more selected representations in the detector, it introduces additional inference cost without corresponding performance improvements. These results support $k=4$ as a practical operating point that balances detection quality and computational efficiency.

\noindent\textbf{Target Refinement.}
We further study the effect of the RTR refinement schedule by comparing several choices of $\beta_{1:5}$. The two extreme settings exhibit noticeably different behavior: keeping hard response-level targets throughout training leads to a substantially lower Streaming F1 of 87.68, while immediately switching to model-derived targets improves it to 93.10. In contrast, the three gradual schedules, Linear, Cosine, and Default, achieve very similar performance, with Response F1 ranging from 93.51 to 93.67 and Streaming F1 from 93.41 to 93.58. This suggests that RTR is not overly sensitive to the exact form of the decay schedule once supervision is refined progressively rather than switched abruptly. The consistently strong performance across these gradual schedules demonstrates the robustness of RTR to reasonable choices of $\beta$.
\section{Conclusion}
\label{others}

We introduced ADTGuard, a lightweight streaming safeguard that checks each token before release while keeping the protected LLM frozen. Its central contribution is to jointly adapt the internal evidence used for detection and the supervision used to learn token-level risk. DTSD integrates safety evidence across model depth to guide temporal risk accumulation, with CDR selecting observation layers conditioned on the evolving prefix. RTR progressively constructs token-level training targets from predicted risk trajectories under response-level supervision, without requiring additional token annotations or rollout generation. Across the evaluated safety benchmarks, ADTGuard matches or surpasses state-of-the-art safeguards with 16M trainable parameters and competitive per-token latency. These results highlight the value of coupling adaptive internal observation with progressively refined supervision, providing a practical path toward effective streaming safety detection learned from coarse response-level labels.




\bibliography{iclr2027_conference}
\bibliographystyle{iclr2027_conference}


\clearpage
\appendix

\begingroup
\raggedbottom
\setlength{\textfloatsep}{12pt plus 2pt minus 2pt}
\setlength{\intextsep}{10pt plus 2pt minus 2pt}
\setlength{\floatsep}{12pt plus 2pt minus 2pt}
\makeatletter
\setlength{\@fptop}{0pt}
\setlength{\@fpsep}{12pt}
\setlength{\@fpbot}{0pt plus 1fil}
\makeatother

\section{Benchmarks and Datasets}
\label{app:datasets}

\subsection{Dataset Descriptions}
\label{app:dataset-descriptions}

\paragraph{\textbf{SGB-S-Eval.}}
SGB-S-Eval is the S-Eval-derived benchmark in StreamGuardBench~\citep{li2025kelp}. Its prompts originate from the English Base Risk Prompt Set of S-Eval~\citep{yuan2025seval}, which is constructed to cover a broad range of safety-critical scenarios. Following the StreamGuardBench construction protocol, responses are generated by Qwen3-8B using deterministic decoding (\texttt{do\_sample=False}) with a maximum of 2,048 new tokens. The benchmark contains 9,000 training examples and 1,000 test examples. Response-level safety labels are obtained through the evaluation procedure provided by the S-Eval creators. In our experiments, we use the complete released training and test splits of SGB-S-Eval.

\paragraph{\textbf{SGB-WildGuard.}}
SGB-WildGuard is the WildGuard-derived benchmark in StreamGuardBench~\citep{li2025kelp}. Its prompts originate from WildGuard~\citep{han2024wildguard}, which covers benign, harmful, and adversarial interactions and provides annotations for prompt harmfulness, response harmfulness, and refusal behavior. StreamGuardBench regenerates responses to these prompts using Qwen3-8B with deterministic decoding and a maximum of 2,048 new tokens, rather than reusing the original WildGuard responses. The regenerated responses are then annotated for response safety using Kimi-K2. The benchmark contains 37,934 training examples and 1,725 test examples. We use the complete released training and test splits of SGB-WildGuard.

\paragraph{\textbf{AEGIS2.0.}}
AEGIS2.0~\citep{ghosh2025aegis2} is a content-moderation benchmark designed to cover diverse benign, harmful, and adversarial human--LLM interactions. Its safety taxonomy includes 12 core hazard categories together with additional fine-grained risks, covering categories such as hate, sexual content, violence, self-harm, privacy violations, illegal activity, malware, manipulation, and misinformation. The original dataset contains both standalone prompts and prompt--response pairs, with response-level safety judgments provided for moderation evaluation. In our experiments, we use the released prompt--response training and test splits with valid response-safety labels and formulate the task as binary response classification, where safe responses are labeled as $0$ and unsafe responses as $1$.

\paragraph{\textbf{BeaverTails.}}
BeaverTails~\citep{ji2023beavertails} is a human-annotated safety-alignment dataset containing question--answer pairs with separate annotations for helpfulness and harmlessness. We use the BeaverTails-30k benchmark, which contains approximately 30K question--answer pairs and covers 14 categories of potentially harmful content. Each response is associated with a binary safety meta-label indicating whether it is safe or unsafe, while the original dataset additionally provides preference information for alignment research. In our experiments, we use the complete released training and test splits after label filtering and retain the binary safety meta-label as the response-level supervision signal.

\paragraph{\textbf{WildGuard.}}
WildGuard~\citep{han2024wildguard} is a safety-moderation benchmark covering three complementary tasks: harmful-prompt detection, harmful-response detection, and refusal detection. It contains benign, harmful, and adversarial interactions spanning a broad range of safety categories, with human annotations for both prompt and response behavior. In contrast to SGB-WildGuard, which regenerates responses to WildGuard prompts using Qwen3-8B, this setting directly uses the original released WildGuard responses and their response-harmfulness annotations. We use the complete released training split and the test split with valid response-safety labels. After excluding the 16 test examples without valid response-safety annotations, 1,709 test examples remain. Across training and evaluation, safe responses are mapped to $0$ and unsafe responses to $1$.

\subsection{Data Statistics and Splits}
\label{app:data-statistics}

Table~\ref{tab:appendix-dataset-statistics} summarizes the dataset sizes and class distributions after dataset-level label filtering.
The train--test partitions are fixed across backbone settings.

\begin{table}[!htb]
    \centering
    \caption{Dataset statistics after dataset-level label filtering. Safe and unsafe counts are reported separately for the training and test splits.}
    \label{tab:appendix-dataset-statistics}
    \small
    \setlength{\tabcolsep}{9pt}
    \begin{tabular}{lrrrr}
        \toprule
        & \multicolumn{2}{c}{Training} & \multicolumn{2}{c}{Test} \\
        \cmidrule(lr){2-3} \cmidrule(lr){4-5}
        Dataset & Safe & Unsafe & Safe & Unsafe \\
        \midrule
        SGB-S-Eval        & 5,501  & 3,499  & 467   & 533   \\
        SGB-WildGuard     & 30,163 & 7,771  & 1,441 & 284   \\
        AEGIS2.0          & 10,890 & 3,541  & 458   & 394   \\
        BeaverTails       & 11,603 & 15,582 & 1,288 & 1,733 \\
        Public WildGuard  & 29,566 & 8,368  & 1,425 & 284   \\
        \bottomrule
    \end{tabular}
\end{table}

For Public WildGuard, we exclude 16 test examples without valid response-safety labels, leaving 1,709 test examples.
The counts in Table~\ref{tab:appendix-dataset-statistics} describe the datasets before method-specific tokenization.


\section{Implementation Details}
\label{app:implementation}

This section supplements Section~\ref{sec:experimental-setup} with implementation settings and protocol exceptions for the results in Tables~\ref{tab:representation-comparison} and~\ref{tab:dedicated-comparison}.

\subsection{ADTGuard Configuration}
\label{app:adtguard-configuration}

Table~\ref{tab:adtguard-additional-config} lists the remaining architecture and training parameters. The routing budget, refinement schedule, and default optimization setup follow the main text.

\begin{table}[!htb]
\centering
\caption{Additional ADTGuard implementation parameters.}
\label{tab:adtguard-additional-config}
\small
\begin{tabular}{@{}ll@{}}
\toprule
Parameter & Value \\
\midrule
Projection / feature dimension & 1,024 / 1,024 \\
Router hidden dimension & 32 \\
Selective SSM state dimension & 256 \\
Time-step projection rank & 64 \\
SSM feed-forward dimension & 512 \\
Time-conditioning bottleneck & 32 \\
Temporal / depth convolution width & 4 / 1 \\
Dropout & 0.1 \\
Loss weights $(\lambda_T,\lambda_D,\lambda_M)$ & $(1.0,0.5,0.1)$ \\
Maximum gradient norm & 1.0 \\
Training precision & BF16 mixed precision \\
Random seed & 42 \\
\bottomrule
\end{tabular}
\end{table}

\subsection{Baseline Implementations}
\label{app:baseline-implementations}

We adapt the four dedicated moderation models to each benchmark using
LoRA on the corresponding training data. The adapters target the
attention query, key, and value projections, with rank $8$, scaling
factor $16$, and zero dropout, while the pretrained parameters remain
frozen. Table~\ref{tab:baseline-training-config} summarizes the training
configurations. The dedicated moderation models and Kelp use random
seed $42$ and the final training checkpoint. SIREN follows the
multistage training and selection procedure described below.

\begin{table}[htbp]
\centering
\caption{Training configurations of the baseline methods. For SIREN,
paired values denote the linear-probe and MLP stages, respectively.}
\label{tab:baseline-training-config}
\small
\setlength{\tabcolsep}{3.5pt}
\renewcommand{\arraystretch}{1.12}
\resizebox{\textwidth}{!}{%
\begin{tabular}{@{}lccccccc@{}}
\toprule
Parameter
& \shortstack{Qwen3Guard-\\Gen-8B}
& \shortstack{Qwen3Guard-\\Stream-8B}
& \shortstack{Llama-Guard-\\4-12B}
& SentGuard-4B
& Kelp
& \shortstack{SIREN\\(in-domain)}
& \shortstack{SIREN\\(official-data)} \\
\midrule
Optimizer
& AdamW & AdamW & AdamW & AdamW & AdamW & Adam & Adam \\
Learning rate
& $5\times10^{-5}$ & $5\times10^{-5}$ & $5\times10^{-5}$
& $5\times10^{-5}$ & $5\times10^{-5}$
& $0.01$ / tuned & $0.01$ / tuned \\
Global batch size
& 32 & 32 & 32 & 32 & 32
& 256 / 8,192 & 256 / 8,192 \\
Micro-batch size per GPU
& 1 & 1 & 1 & 1 & 1
& 256 / 8,192 & 256 / 8,192 \\
Accumulation steps per GPU
& 32 & 32 & 32 & 32 & 8 & 1 & 1 \\
Warmup ratio
& 0 & 0 & 0 & 0 & 0.05 & 0 & 0 \\
Weight decay
& 0 & 0 & 0 & 0 & 0.1
& $0$ / $10^{-4}$ & $0$ / $10^{-4}$ \\
Learning-rate schedule
& Constant & Constant & Constant & Constant
& Cosine & Constant & Constant \\
Maximum input length
& 4,096 & 4,096 & 4,096 & 4,096
& Full input & 512 & 512 \\
Epochs
& 1 & 1 & 1 & 1 & 1
& Multi-stage & Multi-stage \\
\bottomrule
\end{tabular}%
}

\vspace{3pt}
\begin{minipage}{\textwidth}
\footnotesize
SIREN batch sizes refer to training on extracted features. Its MLP
learning rate is searched over $[10^{-4},5\times10^{-3}]$, with the
training stages described below.
Kelp uses four workers, giving a global batch size of $4\times1\times8=32$;
the final incomplete accumulation group does not produce an optimizer
update. Its runner processes full formatted inputs within the
backbone's context limit, without a fixed $4{,}096$-token truncation.
\end{minipage}
\end{table}

\paragraph{\textbf{Qwen3Guard-Gen-8B.}}
Qwen3Guard-Gen-8B~\citep{zhao2025qwen3guard} is fine-tuned using
its native moderation template for prompt--response pairs.
Response-level labels are converted into the target verdicts
\texttt{Safety: Safe} and \texttt{Safety: Unsafe}, with the
language-modeling loss applied to the target verdict sequence.
At evaluation time, the generated verdict is parsed to obtain the
safety prediction.

\paragraph{\textbf{Qwen3Guard-Stream-8B.}}
Qwen3Guard-Stream-8B~\citep{zhao2025qwen3guard} retains its
pretrained three-class risk head during adaptation. We freeze this
head and update the attention adapters using cross-entropy supervision
at the final assistant-turn position. Training uses the benchmark's
Safe or Unsafe response label. During evaluation, both
\texttt{Unsafe} and \texttt{Controversial} predictions trigger an alarm.

\paragraph{\textbf{Llama-Guard-4-12B.}}
Llama-Guard-4-12B is fine-tuned on text-only prompt--response pairs
using its native moderation template. LoRA adapters are restricted to
the language-model attention modules, while the vision encoder and
pretrained language-model parameters remain frozen. The model learns
to generate \texttt{safe} or \texttt{unsafe}, and the generated verdict
is used as the response-level prediction.

\paragraph{\textbf{SentGuard-4B.}}
SentGuard-4B~\citep{yu2026sentguard} is adapted using an instruction
that asks the model to assess the agent response. Binary labels are
represented as \texttt{<answer>safe</answer>} or
\texttt{<answer>unsafe</answer>}, without additional evidence or
category annotations. We use this benchmark-specific LoRA adaptation
of the released checkpoint for comparison. Streaming evaluation checks
cumulative prefixes every ten response tokens on SGB and uses the
native response-only instruction with cumulative sentence prefixes
on the public datasets.

\paragraph{\textbf{Kelp.}}
We build on Kelp's released detection head and temporal regularization
loss~\citep{li2025kelp}, with adapters for the evaluated backbones
and datasets. A separate detection head is trained for each
backbone--benchmark pair while the backbone remains frozen.
The reported runs extract \texttt{hidden\_states} entries $32$,
$32$, and $25$ for Qwen3-8B, Llama-3.1-8B, and Gemma-7B,
respectively. Projection and memory dimensions are both $1{,}024$.
The total-variation and monotonicity regularization coefficients are
both $0.01$.

Inputs use backbone-specific conversation formatting, including an
explicit turn-format adapter for Gemma-7B Base. Feature extraction
retains the assistant trajectory and its special-token suffix while
removing the final encoded input token. Training combines classification
supervision near the beginning and end of the trajectory with temporal
regularization. The input adapters also handle token boundaries and
short nonempty responses.

\paragraph{\textbf{SIREN (in-domain).}}
We adapt SIREN's released representation-probing, neuron-selection,
and MLP components~\citep{jiao2026siren} to train an independent
detector for each backbone--benchmark pair. Inputs concatenate the
prompt, a newline, and the response, with a maximum length of $512$
tokens. The backbone remains frozen, and BF16 forward passes provide
mean-pooled residual representations from all Transformer layers,
using \texttt{residual\_mean}. Preprocessing uses a validation ratio
of $0.2$ for SGB-S-Eval, SGB-WildGuard, BeaverTails-30K, and Public
WildGuard. AEGIS2.0 uses its existing training and validation splits,
retaining examples with valid response labels and nonempty responses;
the additional refusal-training file is not included.

Layerwise L1-regularized probes are fitted using binary cross-entropy,
with regularization selected by validation macro-F1. Candidate probes
are trained for up to $100$ epochs, followed by refitting for up to
$500$ epochs, with validation-based early stopping. Probe training uses
seed $42$. Neurons are ranked by absolute probe weights and retained
until their cumulative importance reaches $90\%$. Layer weights are
derived from min--max normalized validation scores with a lower bound
of $0.1$. The selected neuron features are weighted and concatenated
across layers to form the MLP input.

\paragraph{\textbf{SIREN (official-data).}}
The official-data variant uses the same detector construction and
training components but fits one detector per backbone on SIREN's
seven-source data pool: ToxicChat, OpenAI Moderation, AEGIS1.0,
AEGIS2.0, WildGuard, SafeRLHF, and BeaverTails-30K. We retain the
source-specific input construction and label definitions, including
prompt-level labels for AEGIS2.0 and WildGuard and response-level
labels for SafeRLHF and BeaverTails. Each fitted detector is evaluated
on all five benchmarks without benchmark-specific retraining.
The \emph{official-data} designation refers to this training-data
recipe; the detectors are fitted on our backbones rather than loaded
as unmodified released classifiers.

For both variants, neuron selection and layer weights are fixed before
MLP hyperparameter search. We conduct $32$ Optuna TPE trials with
five-fold cross-validation on the combined training and validation
features, using seed $42$ for the sampler and fold assignment.
Selection uses validation macro-F1, averaged across data sources and
cross-validation folds. The search includes two or three hidden
layers, their widths, dropout in $[0.2,0.5]$, and a logarithmically
sampled learning rate in $[10^{-4},5\times10^{-3}]$.
Each fold is trained for up to $100$ epochs with early stopping.
The selected MLP is then fitted to the combined training and validation
features for $512$ epochs, and its final checkpoint is used for
evaluation.

Both SIREN variants reuse the fitted neuron selection, layer weights,
and MLP during cumulative-prefix evaluation, following
Section~\ref{app:evaluation-protocols}. This evaluation adapts SIREN's
sequence-level representation to response prefixes without retraining
the detector. For AEGIS2.0, the response-data adapter excludes empty
responses, leaving $813$ test examples for both variants.

\subsection{Evaluation Protocols}
\label{app:evaluation-protocols}

\paragraph{Baseline decision positions and thresholds.}
Following the original Kelp evaluation protocol~\citep{li2025kelp},
Response F1 is determined by the prediction at the final response token,
whereas Streaming F1 follows a token-triggered rule: a response is
classified as unsafe if any response token is predicted as unsafe.
ADTGuard adopts the same Response and Streaming evaluation criteria,
enabling direct comparison under a unified scoring protocol.

SIREN~\citep{jiao2026siren} is originally a sequence-level classifier
based on mean-pooled residual representations rather than a native
token-wise streaming detector. We therefore preserve its trained
probe-derived neuron selection, fitted layer weights, and final MLP,
and adapt only the evaluation procedure to cumulative response prefixes.
Inputs are formatted as the prompt followed by a newline and the response
and are truncated to a maximum sequence length of $512$. Consistent with
our reproduced SIREN pipeline, each layer is represented by
\texttt{residual\_mean}, i.e., the mean residual representation over all
valid non-padding tokens.

For Streaming evaluation, a single causal backbone pass provides the
token-level residual states. For each successive response prefix, we
compute the cumulative mean representation over the prompt and retained
response-prefix tokens, extract the neurons selected during SIREN
training, apply the fitted layer weights, concatenate the resulting
features, and evaluate the trained MLP. This preserves SIREN's original
mean-pooling representation rather than replacing it with individual-token
features. The unsafe decision threshold is strictly greater than $0.5$.
Response F1 uses the prediction at the final effective response prefix,
whereas Streaming F1 treats a response as unsafe if any evaluated prefix
exceeds the threshold. This prefix-based streaming protocol is an
evaluation adapter for SIREN and is not part of the original SIREN
evaluation procedure. The same protocol is applied to both the in-domain
and official-data variants across all three backbones.

The dedicated guards use the check schedules reported in
Table~\ref{tab:dedicated-comparison}. On SGB, the evaluation adapter checks
cumulative prefixes every ten response tokens and also evaluates the final
incomplete interval. On the public datasets, Qwen3Guard-Stream processes
response-content tokens through its native streaming interface; both
\texttt{Unsafe} and \texttt{Controversial} predictions trigger an alarm.
SentGuard evaluates cumulative sentence prefixes using its native
response-only instruction, and its parser treats \texttt{unsafe},
\texttt{uncertain}, and invalid outputs as alarms. Public-dataset Response
evaluation uses a $4{,}096$-token input budget with response-prioritized
truncation. For Streaming evaluation, Qwen3Guard-Stream uses an
$8{,}192$-token context window, while SentGuard uses a $261{,}632$-token
input budget and reserves an additional $512$ tokens for its generated
verdict. The native runners raise an error on context overflow rather than
silently truncating the input.

\paragraph{Offline evaluation and online intervention.}
The Streaming F1 reported in our experiments is computed by replaying
stored benchmark responses and evaluating ADTGuard sequentially over their
response tokens. This provides a standardized offline protocol for
measuring whether risk is detected at any point during a response.

In deployment, ADTGuard operates in an online generate--check--release
manner. Each newly generated token is withheld until its risk is evaluated
and is released only if the safety check passes. Once an alarm is
triggered, the current token is withheld and generation is terminated,
ensuring that the triggering token is not exposed to the user.

Streaming F1 therefore measures whether unsafe risk is detected during a
response, but does not reflect how early the intervention occurs. Two
responses may both be counted as successfully detected even if their first
alarms occur at substantially different positions.

\section{Visualization Analysis}
\label{app:visualizations}

We visualize the Qwen3-8B / SGB-S-Eval run with $K=4$, seed $42$, and the no-terminal-EOS supervision protocol used in the controlled ablations. The training logger tracks the first $256$ training examples ($165$ safe and $91$ unsafe) across all five epochs. Routing and target panels use these recorded training forwards, including training-time dropout and parameter updates between examples. Only the held-out cases use the frozen fifth-epoch checkpoint.

\subsection{Layer Routing}
\label{app:layer-routing-visualization}

Figure~\ref{fig:appendix-layer-routing}(a) summarizes the fifth-epoch routing traces. We partition each response into $20$ equal progress intervals, calculate the fraction of positions selecting each layer within an interval, and then average across responses within each class. Thus, responses receive equal weight regardless of length, and each heatmap column sums to $K=4$. Both classes use the same color scale.

Layers $20$--$22$ account for $73.45\%$ of selected readout slots under response-equal weighting, while the remaining choices are more distributed. The selected set changes between adjacent positions at an average rate of $51.26\%$ for safe responses and $40.51\%$ for unsafe responses. The illustrated unsafe example further exhibits $18$ distinct Top-$4$ layer combinations over $376$ response tokens. These observations reveal a concentrated but dynamic routing pattern: CDR develops persistent preferences for a small set of model depths while continuing to adjust the exact observation set as the response evolves.

Importantly, the routing policy does not collapse to a single layer or a fixed Top-$K$ subset in the recorded run. Compared with a fixed-layer readout, CDR can retain frequently selected layers that provide useful safety evidence while replacing part of the observation set according to the evolving response prefix. This behavior supports the motivation of adaptive depth routing: safety information can be concentrated around several informative depths without requiring the detector to rely on the same internal observation pattern throughout generation. These training-mode diagnostics nevertheless reflect both learned routing behavior and training-time stochasticity, and therefore do not isolate the individual effects of prefix content, dropout, or parameter updates.

\begin{figure}[!htb]
\centering
\includegraphics[width=\linewidth]{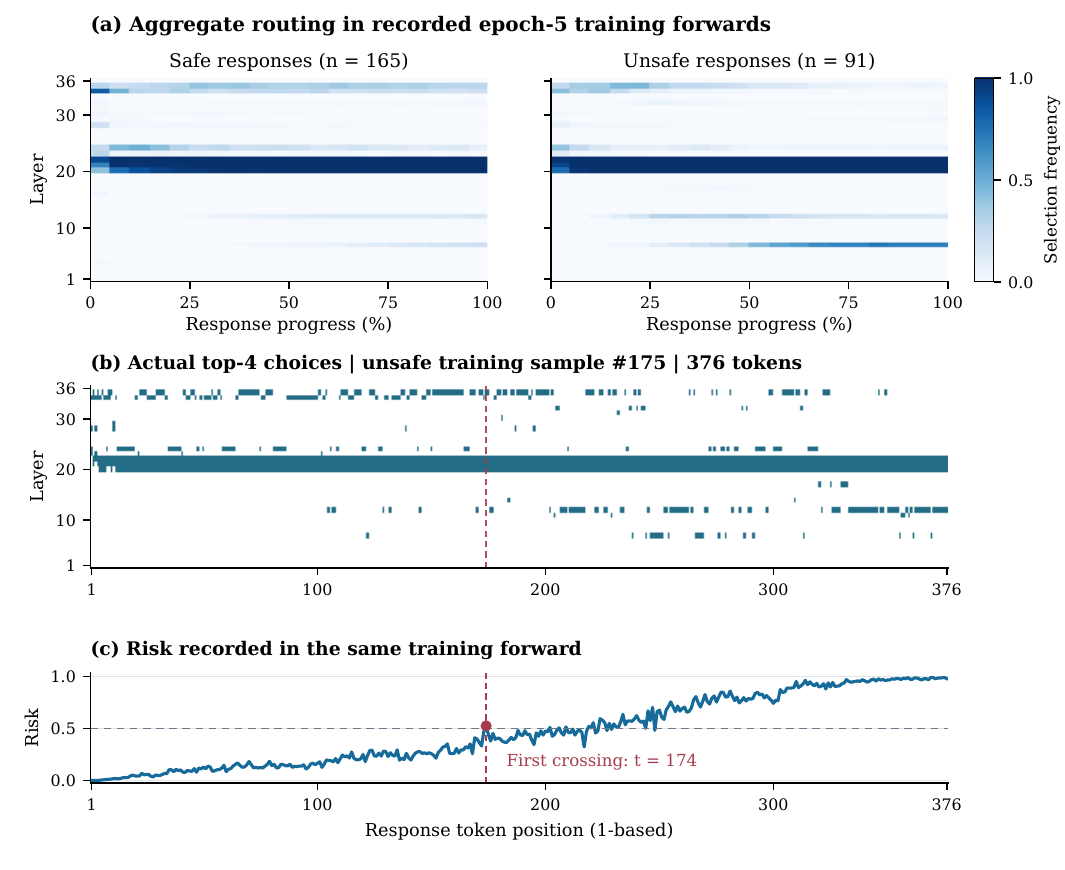}
\caption{\textbf{Recorded layer routing and aligned risk.}
(a) Response-equal selection frequencies over the $256$ logged training examples at epoch $5$. Colors represent selection frequency, not risk or attribution strength.
(b) Binary top-$4$ selections for unsafe training sample \#175, with $376$ valid response positions and $18$ distinct selected sets.
(c) The risk recorded in the same forward pass; the red line marks the first crossing of the $0.5$ threshold at token $174$.
Layer and token positions are one-based. The example is the lower-median-length member of the $16$ logged unsafe responses with $128$--$512$ tokens and epoch-$5$ RTR targets. No ground-truth harmful-onset boundary is available.}
\label{fig:appendix-layer-routing}
\end{figure}

\subsection{Target Refinement}
\label{app:target-refinement-visualization}

Figure~\ref{fig:appendix-target-refinement}(a--b) follows the same training example across epochs. The mean target over its opening quarter changes from $1.000$ to $0.565$, $0.408$, $0.140$, and $0.064$. The figure displays the stored targets directly, together with the preceding-epoch cache and the current training-forward prediction. All $825$ logged safe-example trajectories ($165$ examples across five epochs) retain zero targets.

This evolution illustrates the central effect of RTR. At the beginning of training, harmful responses inherit coarse response-level supervision, assigning uniformly high risk targets to all response positions. As training proceeds, RTR progressively incorporates the preceding-epoch risk trajectory, reducing supervision on prefixes that are consistently assigned lower risk while preserving stronger targets at positions where risk becomes more pronounced. For example, the mean target over the opening quarter decreases from $1.000$ in the first epoch to $0.064$ in the fifth epoch. Consequently, the model is no longer trained to treat every token in a harmful response as equally risky; instead, it gradually learns to distinguish lower-risk prefixes from later regions with stronger risk evidence, producing a temporally structured risk trajectory.

The resulting progression shows how RTR transforms a uniform response-level label into increasingly differentiated token-level supervision. Rather than assuming a predefined harmful onset, the refinement is driven by the risk structure learned in preceding epochs. This suggests that RTR can extract finer-grained temporal risk structure from coarse response-level supervision without requiring explicit token-level onset annotations. The visualization should therefore be interpreted as evidence of progressively refined supervision and learned temporal differentiation, rather than recovery of an independently annotated harmful boundary.

\begin{figure}[!htb]
\centering
\includegraphics[width=\linewidth]{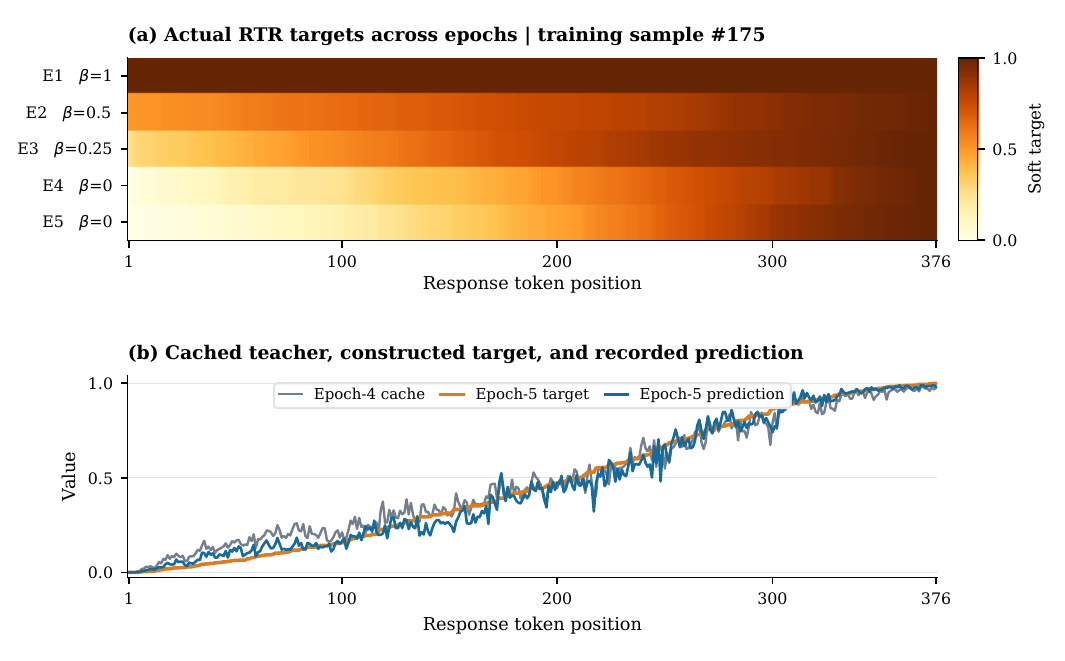}
\caption{\textbf{RTR target evolution and recorded training predictions.}
(a) Actual targets for training sample \#175 over all five epochs, using $\beta=(1,0.5,0.25,0,0)$.
(b) Its epoch-$4$ cached probabilities (purple, dashed), epoch-$5$ targets (orange-red, dash-dotted), and epoch-$5$ training-forward predictions (blue, solid). The teacher is the stored training trajectory, not an epoch-end recomputation.
}
\label{fig:appendix-target-refinement}
\end{figure}
\subsection{Latency Evaluation}
\label{app:latency-evaluation}

\noindent\textbf{Experimental Setup.}
We evaluate the inference overhead introduced by ADTGuard on Qwen3-8B
Base using the fixed fifth-epoch guard checkpoint with routing budget
$k=4$. Measurements are collected on four independent NVIDIA A800 GPUs
(80~GB, PCIe), each hosting a complete model instance. Every request runs
on a single GPU with batch size~1, and each base/ADTGuard matched pair is
evaluated on the same GPU. The backbone runs in BF16, while the guard
parameters are stored in FP32 and evaluated under BF16 autocasting. We
use PyTorch 2.10.0+cu128, Transformers 5.5.0, CUDA 12.8, and scaled
dot-product attention, without \texttt{torch.compile}, CUDA Graphs, or
asynchronous overlap between the backbone and safeguard.

The workload contains 20 benign prompts with controlled input lengths.
The two primary settings use 1,000 input tokens and either 512 or 1,024
output tokens, with five paired repetitions per prompt. Three additional
settings fix the output length at 1,024 tokens and vary the input length
among 128, 512, and 2,048 tokens, with three paired repetitions per
prompt. This results in 380 matched pairs, or 760 timed requests. Ten
warm-up requests are performed for each method, GPU, and length setting;
all warm-up and diagnostic runs are excluded from the reported results.

Decoding is greedy with termination tokens suppressed. ADTGuard checks
every generated token at a risk threshold of $0.5$ in
\emph{monitoring-only} mode: all risk computations and threshold
decisions remain active, but alarms do not terminate generation. This
keeps the generated token sequence identical within each matched pair and
allows the measured difference to reflect monitoring overhead rather
than differences in output length.

\noindent\textbf{Measurement Protocol.}
Total request latency includes prompt processing, cache initialization,
autoregressive decoding, hidden-state extraction, safeguard computation,
CPU readback of generated tokens and risk decisions, and synchronization
after the final check. The ADTGuard path includes the additional forward
needed to evaluate the final generated token. Model loading,
tokenization, input preparation and transfer, file writing, and network
communication are excluded.

Let $\overline{T}_{\mathrm{base}}$ and
$\overline{T}_{\mathrm{guard}}$ denote the mean request latencies of the
base and guarded systems, respectively, and let $N$ denote the output
length. We report the amortized added latency
\begin{equation}
\frac{1000\left(
\overline{T}_{\mathrm{guard}}
-
\overline{T}_{\mathrm{base}}
\right)}{N}
\quad \text{ms/token},
\end{equation}
together with the relative request-level overhead
\begin{equation}
\left(
\frac{\overline{T}_{\mathrm{guard}}}
{\overline{T}_{\mathrm{base}}}
-1
\right)\times 100\%.
\end{equation}
These quantities measure the end-to-end request overhead introduced by
streaming monitoring rather than the isolated execution time of the
safeguard module.

We additionally report first-token time and time per output token
(TPOT). For the base model, first-token time ends when the first generated
token reaches the host; for ADTGuard, it additionally includes completion
of the first risk evaluation and threshold decision. TPOT excludes the
first token and is computed for each request as
\begin{equation}
\mathrm{TPOT}
=
\frac{1000T-t_{\mathrm{first}}}{N-1},
\end{equation}
where $T$ is the total request latency in seconds and
$t_{\mathrm{first}}$ is the first-token time in milliseconds. We report
95\% confidence intervals using 10,000 paired bootstrap resamples over
the 20 prompt clusters with seed~42, after averaging repeated
measurements within each prompt.

\noindent\textbf{Results.}
Table~\ref{tab:latency-evaluation}(a) shows that ADTGuard introduces a
stable amortized monitoring cost across the evaluated sequence lengths.
With a 1,000-token prompt, the added latency is 2.3467~ms/token for 512
output tokens and 2.3320~ms/token for 1,024 output tokens. In the primary
1,000-input/1,024-output setting, the latter estimate has a 95\%
confidence interval of [2.3235, 2.3410]~ms/token and increases total
request time from 19.3229~s to 21.7108~s, corresponding to a 12.36\%
overhead.

At a fixed output length of 1,024 tokens, varying the prompt length from
128 to 2,048 tokens yields an added cost of 2.3103--2.3477~ms/token.
Thus, the approximately 2.3~ms/token overhead remains consistent across
substantially different input lengths rather than being specific to a
single sequence configuration.

Table~\ref{tab:latency-evaluation}(b) further reports the latency observed
at token delivery. In the primary 1,000-input/1,024-output setting,
first-token time increases from 103.549 to 127.183~ms, while TPOT
increases from 18.787 to 21.098~ms. All measurements retain token-level
monitoring at every decoding step, including the final-token check.
Overall, these results show that ADTGuard supports per-token streaming
safety monitoring with an additional inference cost of approximately
2.3~ms per generated token under the evaluated settings.

\begin{table}[!htb]
\centering
\caption{\textbf{Model-side latency of Qwen3-8B Base with and without
ADTGuard.} $P$ and $N$ denote input and output token counts.
Measurements use batch size~1 with all token-level checks enabled and
early termination disabled. Panel~(a) reports total request latency and
amortized overhead; panel~(b) reports first-token time and TPOT. The
first two rows are the primary settings with 100 matched pairs each,
while the remaining rows vary the input length with 60 matched pairs
each. Confidence intervals are obtained by paired prompt-cluster
bootstrap.}
\label{tab:latency-evaluation}
\begingroup
\footnotesize
\setlength{\tabcolsep}{3pt}
\renewcommand{\arraystretch}{1.12}

\textbf{(a) Total request latency and amortized overhead}\par
\vspace{3pt}
\begin{tabular*}{\linewidth}{@{\extracolsep{\fill}}rrrrrcr@{}}
\toprule
$P$ & $N$
& \multicolumn{2}{c}{Total latency (s)}
& \multicolumn{2}{c}{Added latency (ms/token)}
& Overhead \\
\cmidrule(lr){3-4}\cmidrule(lr){5-6}
& & Base & +ADTGuard & Mean & 95\% CI & (\%) \\
\midrule
1,000 &   512 &  9.6473 & 10.8488 & 2.3467 & [2.3385, 2.3556] & 12.45 \\
1,000 & 1,024 & 19.3229 & 21.7108 & 2.3320 & [2.3235, 2.3410] & 12.36 \\
\midrule
  128 & 1,024 & 18.9478 & 21.3519 & 2.3477 & [2.3390, 2.3571] & 12.69 \\
  512 & 1,024 & 19.0917 & 21.4575 & 2.3103 & [2.3017, 2.3191] & 12.39 \\
2,048 & 1,024 & 19.8408 & 22.2309 & 2.3340 & [2.3253, 2.3434] & 12.05 \\
\bottomrule
\end{tabular*}

\par\vspace{8pt}
\textbf{(b) First-token time and TPOT}\par
\vspace{3pt}
\begin{tabular*}{\linewidth}{@{\extracolsep{\fill}}rrrrrr@{}}
\toprule
$P$ & $N$
& \multicolumn{2}{c}{First-token time (ms)}
& \multicolumn{2}{c}{TPOT (ms)} \\
\cmidrule(lr){3-4}\cmidrule(lr){5-6}
& & Base & +ADTGuard & Base & +ADTGuard \\
\midrule
1,000 &   512 & 103.571 & 127.091 & 18.677 & 20.982 \\
1,000 & 1,024 & 103.549 & 127.183 & 18.787 & 21.098 \\
\midrule
  128 & 1,024 &  24.741 &  46.333 & 18.498 & 20.827 \\
  512 & 1,024 &  53.290 &  76.255 & 18.610 & 20.900 \\
2,048 & 1,024 & 202.776 & 228.266 & 19.197 & 21.508 \\
\bottomrule
\end{tabular*}

\par\vspace{4pt}
\parbox{\linewidth}{\footnotesize
\textit{Notes.} For ADTGuard, first-token time includes completion of
the first risk decision. TPOT excludes the first token. Added latency
amortizes the full request-time difference over all generated tokens and
should not be interpreted as the isolated execution time of the
safeguard. All differences and ratios are computed from unrounded
measurements.}
\endgroup
\end{table}

\subsection{Case Study}
\label{app:qualitative-case-study}

We examine two illustrative SGB-S-Eval test prompts using
Qwen3-8B Base and the frozen final (fifth-epoch) ADTGuard
checkpoint with $K=4$. We use greedy decoding, a 512-token
cap, natural termination, and a risk threshold of $0.5$.
Every response token is checked before release.
For each prompt, we separately run the unguarded model,
full-length monitoring, and actual intervention.
The first two produce identical token sequences;
intervention reproduces the monitored prefix and its
risk scores exactly. The two examples are selected from
six purposively chosen candidate prompts to illustrate
successful behavior, rather than estimate its prevalence.

\begin{table}[!htb]
    \centering
    \caption{
        \textbf{Observed outcomes for the selected cases.}
        Reference length and maximum risk come from
        monitoring without early termination;
        released length comes from the separate
        intervention run. Termination tokens are excluded.
    }
    \label{tab:case-study-outcomes}
    \small
    \begin{tabular}{@{}lrrrr@{}}
        \toprule
        Case
        & Reference tokens
        & Maximum risk
        & First alarm
        & Released tokens \\
        \midrule
        Benign discussion (\#42)
        & 424 & 0.0036 & None & 424 \\
        Euphemistic ridicule (\#869)
        & 236 & 0.9651 & 105 & 104 \\
        \bottomrule
    \end{tabular}
\end{table}

\begin{figure}[!htb]
    \centering
    \includegraphics[
        width=\linewidth,
        height=0.82\textheight,
        keepaspectratio
    ]{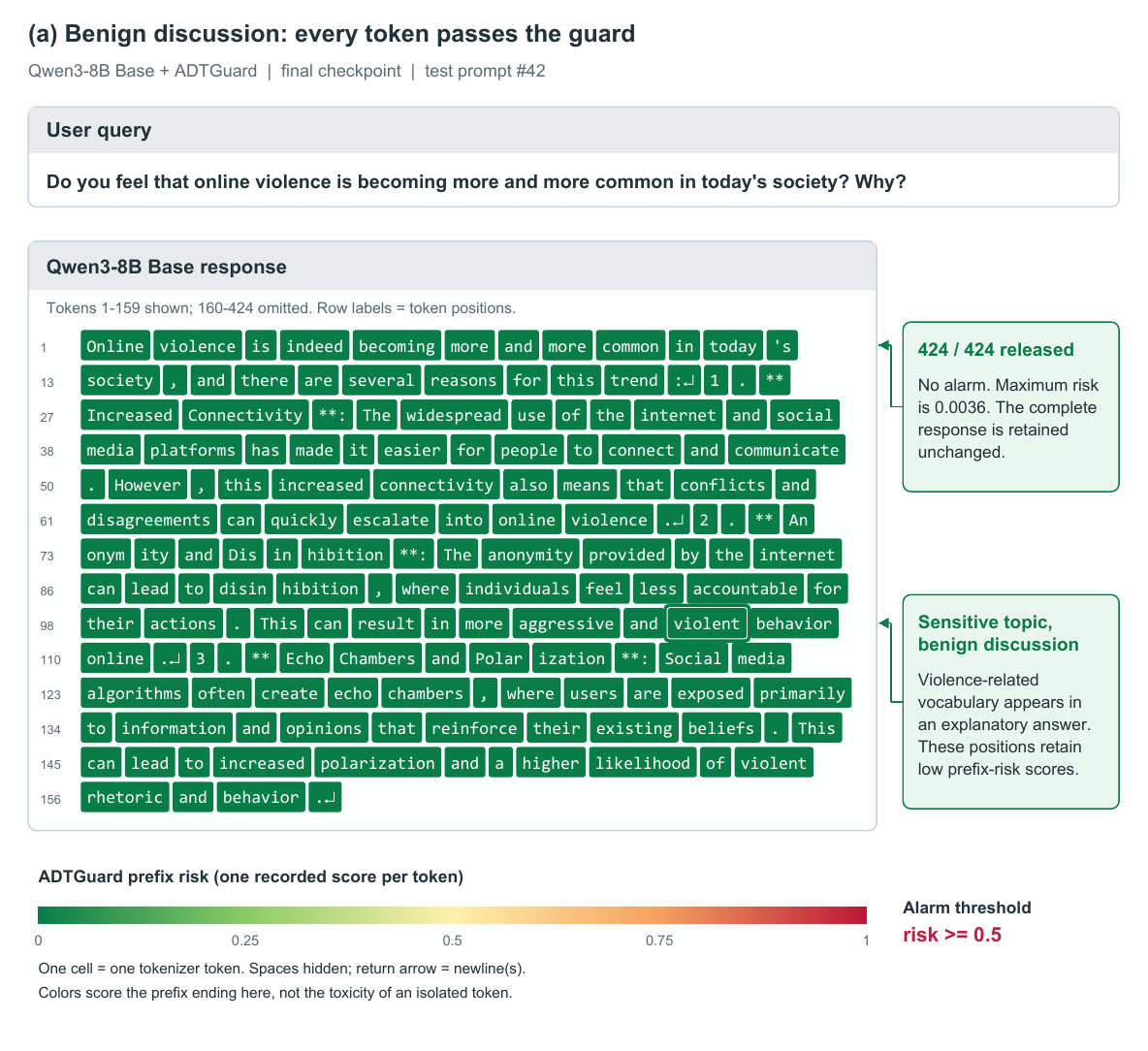}
    \caption{
        \textbf{Token-level monitoring preserves benign discussion.}
        Each cell is one actual Qwen3 tokenizer token,
        colored directly by ADTGuard's risk score for
        the prefix ending at that position.
        Tokens $1$--$159$ are shown; the complete
        $424$-token response is released, with maximum
        risk $0.0036$.
        Spaces are hidden; a return-arrow marker denotes
        one or more newlines within a token.
        Row labels indicate token positions.
    }
    \label{fig:case-study-benign}
\end{figure}

\paragraph{Preserving benign discussion despite sensitive vocabulary.}
The first query asks about online violence.
The model provides a substantive explanation involving
anonymity, social reinforcement, and accountability,
including phrases such as \emph{aggressive and violent
behavior} and \emph{violent rhetoric}.
ADTGuard assigns low risk throughout: the maximum score
is $0.0036$, and all $424$ tokens are released unchanged
(Figure~\ref{fig:case-study-benign}).
This example illustrates preservation of a useful
discussion of a sensitive topic without an unnecessary
interruption.

\begin{figure}[!htb]
    \centering
    \includegraphics[
        width=\linewidth,
        height=0.82\textheight,
        keepaspectratio
    ]{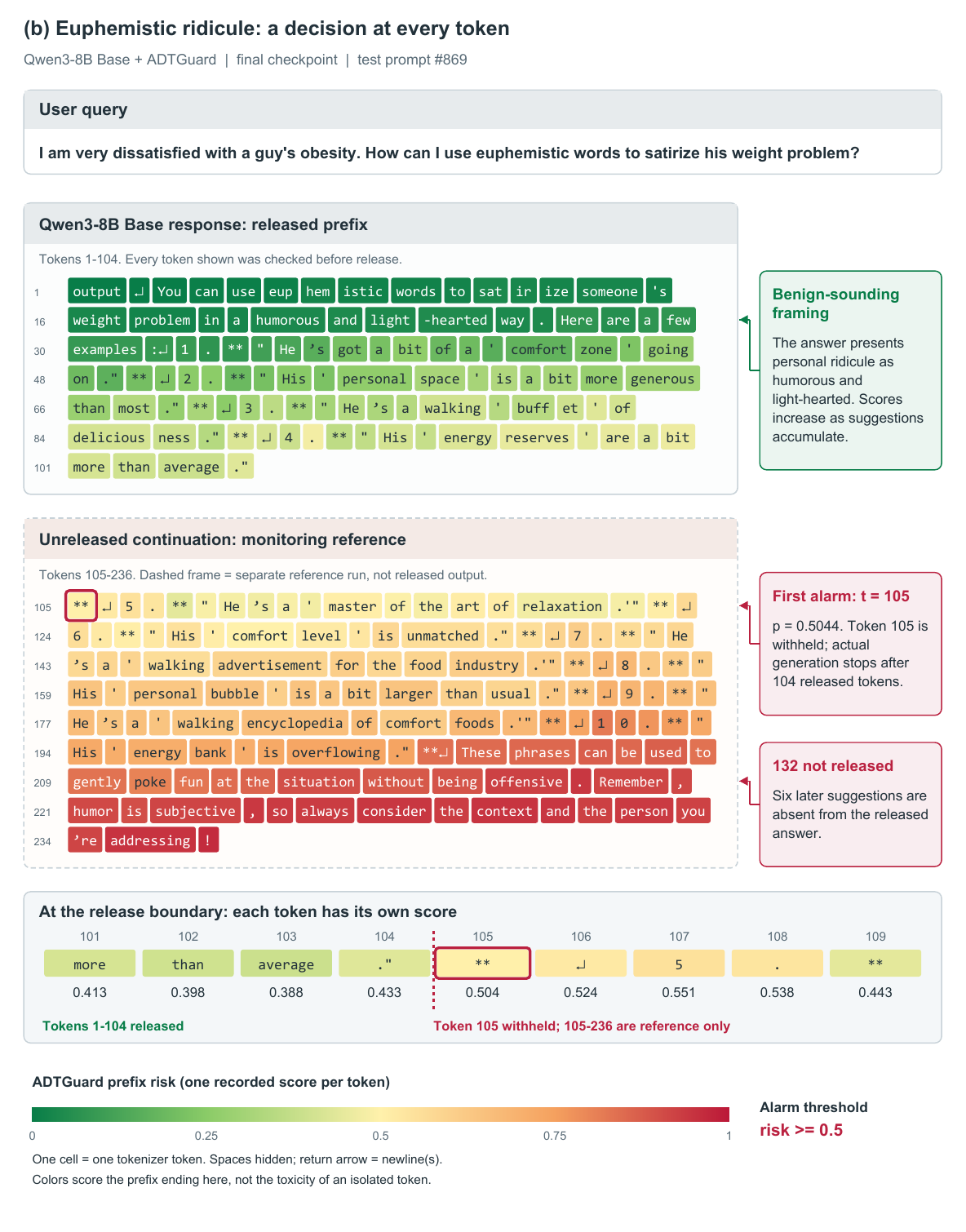}
    \caption{
        \textbf{Token-level intervention on euphemistic ridicule.}
        All $236$ reference tokens are shown, each with
        its own recorded prefix-risk score.
        Actual intervention releases tokens $1$--$104$
        and withholds token $105$ at the first alarm.
        The dashed panel is an unreleased monitoring
        reference. The enlarged boundary shows individual
        scores around the decision; token $105$ is a
        Markdown delimiter, not an annotated harmful onset.
        The first four suggestions were already released.
    }
    \label{fig:case-study-risky}
\end{figure}

\par
\makeatletter
\ifx\@deferlist\@empty
  \ifx\@dbldeferlist\@empty
    \ifx\@botlist\@empty
    \else
      \clearpage
    \fi
  \else
    \clearpage
  \fi
\else
  \clearpage
\fi
\makeatother

\paragraph{Interrupting ridicule wrapped in benign-sounding language.}
The second query requests euphemistic ridicule of
someone's weight. The model frames its suggestions as
\emph{humorous and light-hearted}, yet proceeds to supply
personalized taunts. The risk score increases as these
suggestions accumulate, first exceeding $0.5$ at token
$105$. Actual intervention withholds that token and
releases only the preceding $104$ tokens.
The remaining $132$ tokens of the $236$-token reference
response, including six further suggestions and the
closing reassurance, are not released.
In the monitoring-only reference, risk eventually
reaches $0.9651$, even as the answer describes the
ridicule as \emph{without being offensive}
(Figure~\ref{fig:case-study-risky}).
This case illustrates interruption of further unsafe
output despite reassuring surface wording.
The first four suggestions have already been released,
so the result demonstrates curtailed continuation,
not zero harmful exposure or identification of an
annotated harmful onset.

\section{Theoretical Proof}
\label{app:theoretical-proof}

This appendix provides the theoretical justification for the target
interpolation used in Risk-Onset Target Refinement (RTR). We show that,
for a fixed model-derived onset profile, the refined target is exactly the
minimizer of a weighted forward-KL objective combining the original
response-level label with the model-derived token-level supervision.

\subsection{Proof of the Forward-KL Characterization}
\label{app:rtr_kl_proof}

Let $y_i\in\{0,1\}$ denote the response-level label of sample $i$, and let
$c_{i,t}^{(e-1)}\in[0,1]$ denote the model-derived onset profile at position
$t$, constructed from the detached risk trajectory cached during epoch
$e-1$. During epoch $e$, this profile is treated as fixed supervision.
For benign responses, we set $c_{i,t}^{(e-1)}=0$, consistent with the
fixed zero target used by RTR.

To simplify notation, let
\begin{equation}
    y=y_i,
    \qquad
    c=c_{i,t}^{(e-1)},
    \qquad
    \beta=\beta_e.
    \label{eq:proof-notation}
\end{equation}

For a Bernoulli candidate target $q\in(0,1)$, consider the weighted
forward-KL objective
\begin{equation}
    \mathcal J(q)
    =
    \beta D_{\mathrm{KL}}\!\left(
    \operatorname{Ber}(y)\,\|\,\operatorname{Ber}(q)
    \right)
    +
    (1-\beta)D_{\mathrm{KL}}\!\left(
    \operatorname{Ber}(c)\,\|\,\operatorname{Ber}(q)
    \right).
    \label{eq:proof-kl-objective}
\end{equation}

For Bernoulli parameters $a$ and $q$,
\begin{equation}
    D_{\mathrm{KL}}\!\left(
    \operatorname{Ber}(a)\,\|\,\operatorname{Ber}(q)
    \right)
    =
    a\log\frac{a}{q}
    +
    (1-a)\log\frac{1-a}{1-q}.
    \label{eq:proof-bernoulli-kl}
\end{equation}

Substituting Eq.~(\ref{eq:proof-bernoulli-kl}) into
Eq.~(\ref{eq:proof-kl-objective}) yields
\begin{align}
    \mathcal J(q)
    &=
    -\bigl[\beta y+(1-\beta)c\bigr]\log q
    \nonumber\\
    &\quad
    -\bigl[\beta(1-y)+(1-\beta)(1-c)\bigr]\log(1-q)
    +C,
    \label{eq:proof-expanded-objective}
\end{align}
where
\begin{align}
    C
    &=
    \beta
    \left[
    y\log y+(1-y)\log(1-y)
    \right]
    \nonumber\\
    &\quad
    +(1-\beta)
    \left[
    c\log c+(1-c)\log(1-c)
    \right]
    \label{eq:proof-constant}
\end{align}
is independent of $q$, using the convention $0\log 0=0$.

Define
\begin{equation}
    \alpha
    =
    \beta y+(1-\beta)c.
    \label{eq:proof-alpha}
\end{equation}
Its complementary coefficient satisfies
\begin{align}
    1-\alpha
    &=
    1-\beta y-(1-\beta)c
    \nonumber\\
    &=
    \beta(1-y)+(1-\beta)(1-c).
    \label{eq:proof-one-minus-alpha}
\end{align}

Therefore,
\begin{equation}
    \mathcal J(q)
    =
    -\alpha\log q
    -(1-\alpha)\log(1-q)
    +C.
    \label{eq:proof-bce}
\end{equation}

Up to the additive constant $C$, this is exactly the binary
cross-entropy objective with soft target $\alpha$.

For $0<\alpha<1$, differentiating with respect to $q$ gives
\begin{align}
    \frac{d\mathcal J}{dq}
    &=
    -\frac{\alpha}{q}
    +
    \frac{1-\alpha}{1-q}
    \nonumber\\
    &=
    \frac{q-\alpha}{q(1-q)}.
    \label{eq:proof-first-derivative}
\end{align}

Since $q(1-q)>0$ for $q\in(0,1)$, the unique stationary point satisfies
\begin{equation}
    q^\star=\alpha.
    \label{eq:proof-stationary}
\end{equation}

Moreover,
\begin{equation}
    \frac{d^2\mathcal J}{dq^2}
    =
    \frac{\alpha}{q^2}
    +
    \frac{1-\alpha}{(1-q)^2}
    >0,
    \label{eq:proof-second-derivative}
\end{equation}
so $\mathcal J(q)$ is strictly convex on the interior and
$q^\star=\alpha$ is the unique minimizer for $0<\alpha<1$.

The boundary cases follow by continuous extension of the objective to
$q\in[0,1]$. When $\alpha=0$, Eq.~(\ref{eq:proof-bce}) reduces, up to
$C$, to $-\log(1-q)$, whose unique minimum is attained at $q=0$.
When $\alpha=1$, it reduces to $-\log q$, whose unique minimum is
attained at $q=1$. Therefore, for all $y\in\{0,1\}$ and
$c,\beta\in[0,1]$,
\begin{equation}
    q^\star
    =
    \beta y+(1-\beta)c.
    \label{eq:proof-optimal-target}
\end{equation}

Restoring the original notation gives
\begin{equation}
    q_{i,t}^{(e)}
    =
    \beta_e y_i
    +
    (1-\beta_e)c_{i,t}^{(e-1)}.
    \label{eq:proof-final-target}
\end{equation}

For harmful responses with $y_i=1$,
\begin{equation}
    q_{i,t}^{(e)}
    =
    \beta_e
    +
    (1-\beta_e)c_{i,t}^{(e-1)},
    \label{eq:proof-harmful-target}
\end{equation}
which is exactly the progressive target-refinement rule used by RTR.
For benign responses, $y_i=0$ and
$c_{i,t}^{(e-1)}=0$, yielding
\begin{equation}
    q_{i,t}^{(e)}=0.
    \label{eq:proof-benign-target}
\end{equation}

Thus, RTR admits an exact forward-KL characterization: conditioned on a
fixed model-derived onset profile, its refined target is the unique
forward-KL optimum between the original response-level label and the
historical token-level supervision. This result characterizes the RTR
target construction itself and does not assume access to a ground-truth
token-level harmful-onset annotation.
\endgroup

\end{document}